\documentclass{article} 
\usepackage{iclr2021_conference,times}

\usepackage{amsmath,amsfonts,bm}

\def\eqref#1{equation~\ref{#1}}

\def\1{\bm{1}}

\DeclareMathAlphabet{\mathsfit}{\encodingdefault}{\sfdefault}{m}{sl}
\SetMathAlphabet{\mathsfit}{bold}{\encodingdefault}{\sfdefault}{bx}{n}

\usepackage{hyperref}
\usepackage{url}

\usepackage{multirow}
\usepackage{graphicx}
\usepackage{wrapfig}
\usepackage{caption}
\usepackage{booktabs}

\iclrfinalcopy

\title{Towards Faster and Stabilized GAN Training for High-fidelity Few-shot Image Synthesis}
\author{Bingchen Liu$^{1,2}$, Yizhe Zhu$^2$, Kunpeng Song$^{1,2}$, Ahmed Elgammal$^{1,2}$\\
$^1$Playform - Artrendex Inc., USA \\
$^2$Department of Computer Science, Rutgers University\\
\texttt{\{bingchen.liu,yizhe.zhu,kunpeng.song\}@rutgers.edu} \\
\texttt{elgammal@artrendex.com}
}
\date{September 2020}

\begin{document}

\maketitle

\begin{abstract}
Training Generative Adversarial Networks (GAN) on high-fidelity images usually requires large-scale GPU-clusters and a vast number of training images. In this paper, we study the few-shot image synthesis task for GAN with minimum computing cost. We propose a light-weight GAN structure that gains superior quality on $1024\times1024$ resolution. Notably, the model converges from scratch with just a few hours of training on a single RTX-2080 GPU, and has a consistent performance, even with less than 100 training samples. Two technique designs constitute our work, a skip-layer channel-wise excitation module and a self-supervised discriminator trained as a feature-encoder. With thirteen datasets covering a wide variety of image domains \footnote{The datasets and code are available at: https://github.com/odegeasslbc/FastGAN-pytorch}, we show our model's superior performance compared to the state-of-the-art StyleGAN2, when data and computing budget are limited.
\end{abstract}

\section{Introduction}

The fascinating ability to synthesize images using the state-of-the-art (SOTA) Generative Adversarial Networks (GANs) \citep{goodfellow2014generative} display a great potential of GANs for many intriguing real-life applications, such as image translation, photo editing, and artistic creation. However, expensive computing cost and the vast amount of required training data limit these SOTAs in real applications with only small image sets and low computing budgets.

In real-life scenarios, the available samples to train a GAN can be minimal, such as the medical images of a rare disease, a particular celebrity's portrait set, and a specific artist's artworks. Transfer-learning with a pre-trained model \citep{mo2020freeze,wang2020minegan} is one solution for the lack of training images. Nevertheless, there is no guarantee to find a compatible pre-training dataset. Furthermore, if not, fine-tuning probably leads to even worse performance \citep{zhao2020differentiable}. 

\begin{figure}[h]
\begin{center}
\includegraphics[width=\linewidth]{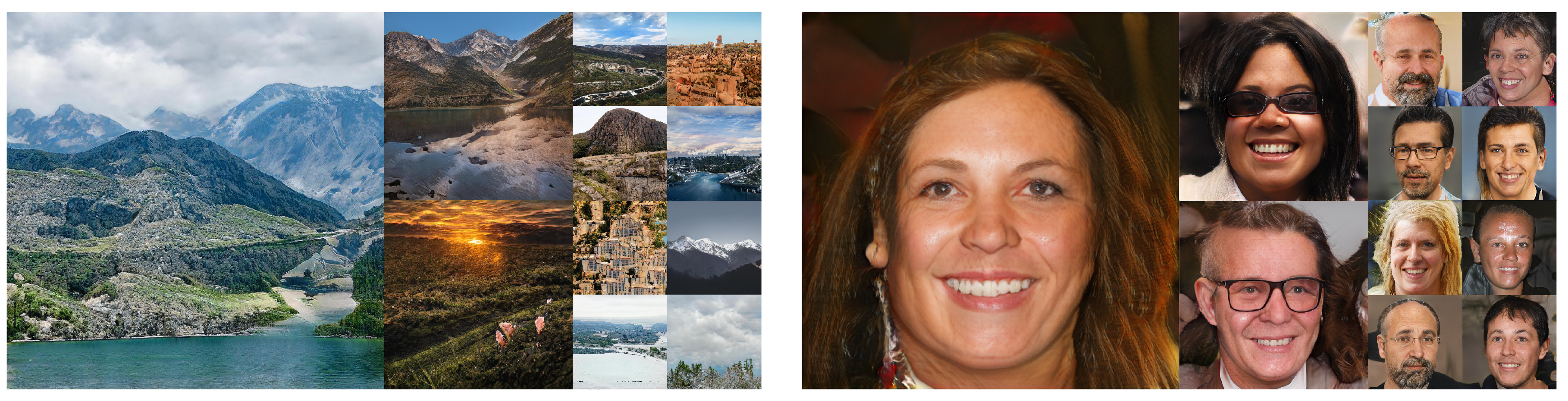}
\end{center}
\caption{\textbf{Synthetic results on $1024^2$ resolution} of our model, trained from scratch on single RTX 2080-Ti GPU, with only 1000 images. Left: 20 hours on Nature photos; Right: 10 hours on FFHQ.}
\label{fig:quality-ours}
\end{figure}

In a recent study, it was highlighted that in art creation applications, most artists prefers to train their models from scratch based on their own images to avoid biases from fine-tuned pre-trained model. Moreover, It was shown that in most cases artists want to train their models with datasets of less than 100 images \citep{elgammal2020artists}. Dynamic data-augmentation \citep{karras2020training,zhao2020differentiable} smooths the gap and stabilizes GAN training with fewer images. However, the computing cost from the SOTA models such as StyleGAN2 \citep{karras2020analyzing} and BigGAN \citep{brock2018large} remain to be high, especially when trained with the image resolution on $1024\times1024$. 

\begin{wrapfigure}{r}{0.5\textwidth} 
\centering
    \includegraphics[width=1\linewidth]{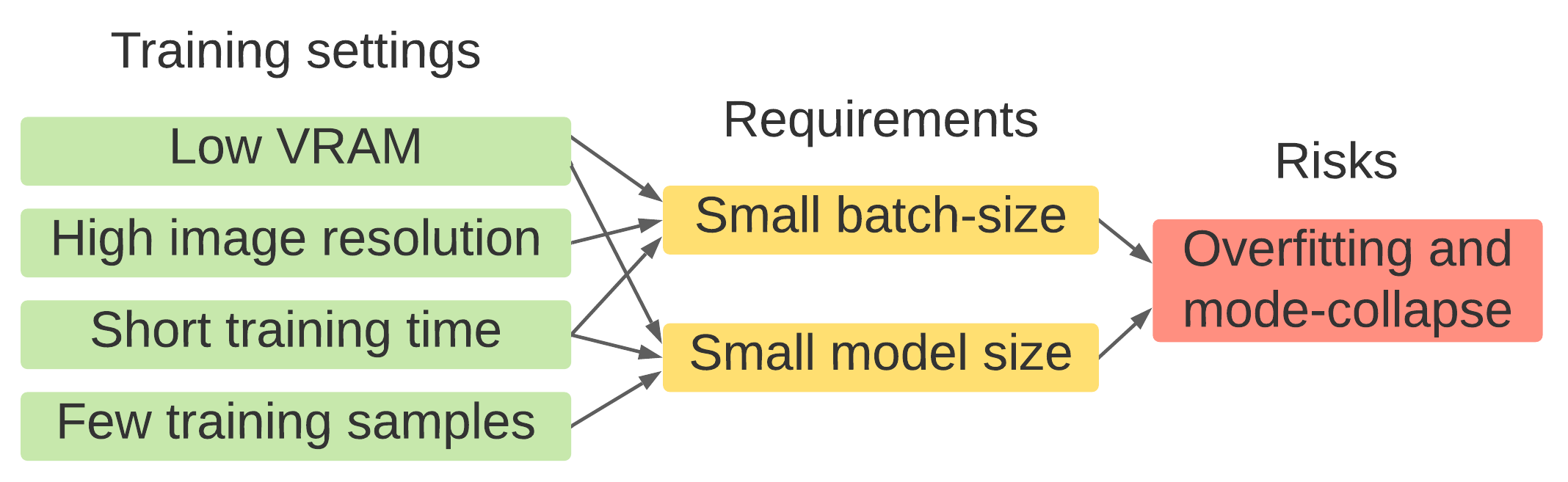}
    \caption{The causes and challenges for training GAN in our studied conditions.}
    \label{fig:challenges}
\end{wrapfigure}

In this paper, our goal is to learn an unconditional GAN on high-resolution images, with low computational cost and few training samples. As summarized in Fig.~\ref{fig:challenges}, these training conditions expose the model to a high risk of  overfitting and mode-collapse~\citep{arjovsky2017towards,zhang2018pa}. To train a GAN given the demanding training conditions, we need a generator $G$ that can learn fast, and a discriminator $D$ that can continuously provide useful signals to train $G$. To address these challenges, we summarize our contribution as:
\begin{itemize}
\item We design the Skip-Layer channel-wise Excitation (SLE) module, which leverages low-scale activations to revise the channel responses on high-scale feature-maps. SLE allows a more robust gradient flow throughout the model weights for faster training. It also leads to an automated learning of a style/content disentanglement like StyleGAN2.  
\item We propose a self-supervised discriminator $D$ trained as a feature-encoder with an extra decoder. We force $D$ to learn a more descriptive feature-map covering more regions from an input image, thus yielding more comprehensive signals to train $G$. We test multiple self-supervision strategies for $D$, among which we show that auto-encoding works the best.
\item We build a computational-efficient GAN model based on the two proposed techniques, and show the model's robustness on multiple high-fidelity datasets, as demonstrated in Fig.~\ref{fig:quality-ours}.
\end{itemize}

\section{Related Works}

\textbf{Speed up the GAN training}: Speeding up the training of GAN has been approached from various perspectives. \citeauthor{ngxande2019depthwisegans} propose to reduce the computing time with depth-wise convolutions. \citeauthor{zhong2020improving} adjust the GAN objective into a min-max-min problem for a shorter optimization path.  \citeauthor{sinha2019small} suggest to prepare each batch of training samples via a core-set selection, leverage the better data preparation for a faster convergence. However, these methods only bring a limited improvement in training speed. Moreover, the synthesis quality is not advanced within the shortened training time.

\textbf{Train GAN on high resolution}: High-resolution training for GAN can be problematic. Firstly, the increased model parameters lead to a more rigid gradient flow to optimize $G$. Secondly, the target distribution formed by the images on $1024\times1024$ resolution is super sparse, making GAN much harder to converge. \cite{denton2015deep,zhang2017stackgan,huang2017stacked,wang2018high,karras2019style,karnewar2019msg,karras2020analyzing,liu2020time} develop the multi-scale GAN structures to alleviate the gradient flow issue, where $G$ outputs images and receives feedback from several resolutions simultaneously. However, all these approaches further increase the computational cost, consuming even more GPU memory and training time. 

\textbf{Stabilize the GAN training}: 
Mode-collapse on $G$ is one of the big challenges when training GANs. And it becomes even more challenging given fewer training samples and a lower computational budget (a smaller batch-size). As $D$ is more likely to be overfitting on the datasets, thus unable to provide meaningful gradients to train $G$ \citep{gulrajani2017improved}.

Prior works tackle the overfitting issue by seeking a good regularization for $D$, including different objectives \citep{arjovsky2017wasserstein,lim2017geometric,tran2017deep}; regularizing the gradients \citep{gulrajani2017improved,mescheder2018training}; normalizing the model weights \citep{miyato2018spectral}; and augmenting the training data \citep{karras2020training,zhao2020differentiable}. However, the effects of these methods degrade fast when the training batch-size is limited, since appropriate batch statistics can hardly be calculated for the regularization (normalization) over the training iterations. 

Meanwhile, self-supervision on $D$ has been shown to be an effective method to stabilize the GAN training as studied in \cite{tran2019self,chen2019self}. However, the auxiliary self-supervision tasks in prior works have limited using scenario and image domain. Moreover, prior works only studied on low resolution images ($32^2$ to $128^2$), and without a computing resource limitation.  

\section{Method}

\begin{figure}[h]
\begin{center}
\includegraphics[width=0.9\linewidth,height=3.9cm]{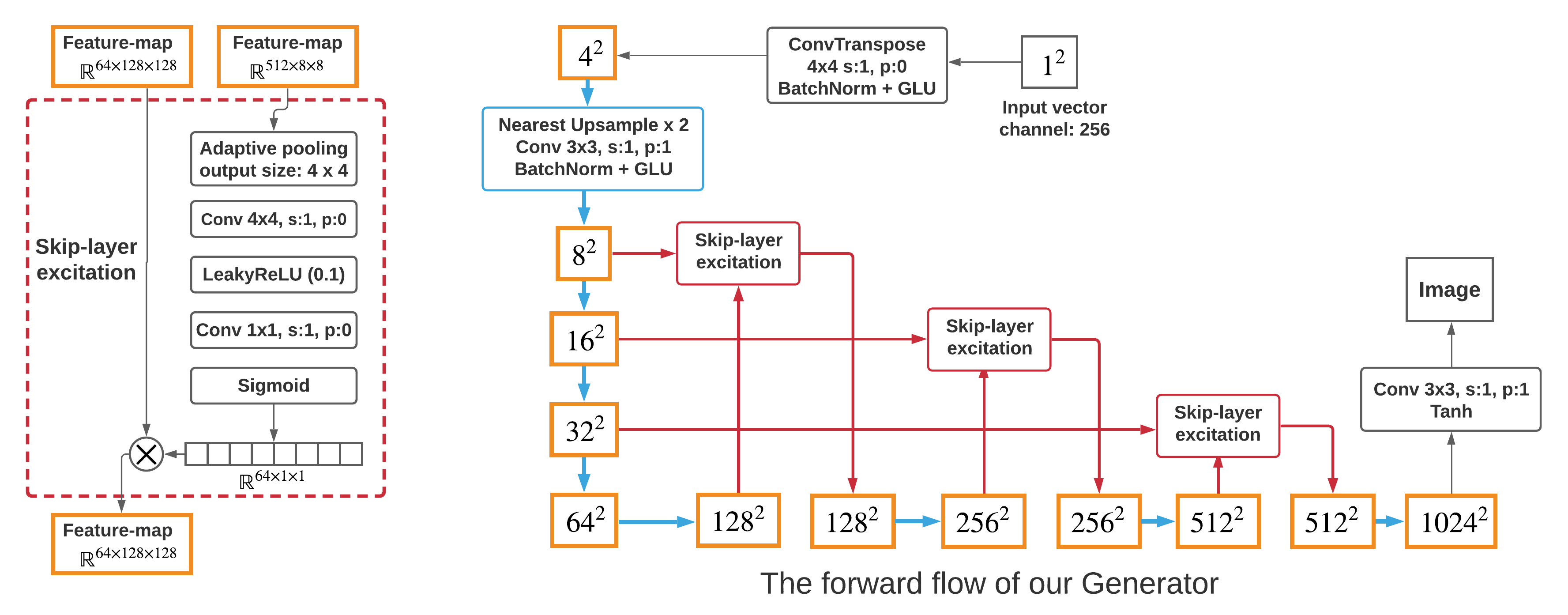}
\end{center}
\caption{The structure of the skip-layer excitation module and the Generator. Yellow boxes represent feature-maps (we show the spatial size and omit the channel number), blue box and blue arrows represent the same up-sampling structure, red box contains the SLE module as illustrated on the left.}
\label{fig:g_overview}
\end{figure}

We adopt a minimalistic design for our model. In particular, we use a single conv-layer on each resolution in $G$, and apply only three (input and output) channels for the conv-layers on the high resolutions ($\geq512\times512$) in both $G$ and $D$. 
Fig.~\ref{fig:g_overview} and Fig.~\ref{fig:d_overview}  illustrate the model structure for our $G$ and $D$, with descriptions of the component layers and forward flow. These structure designs make our GAN much smaller than SOTA models and substantially faster to train. Meanwhile, our model remains robust on small datasets due to its compact size with the two proposed techniques.

\subsection{Skip-Layer channel-wise Excitation}

For synthesizing higher resolution images, the generator $ G $ inevitably needs to become deeper, with more conv-layers, in concert with the up-sampling needs. A deeper model with more convolution layers leads to a longer training time of GAN, due to the increased number of model parameters and a weaker gradient flow through $G$ \citep{zhang2017stackgan,karras2017progressive,karnewar2019msg}. To better train a deep model, \citeauthor{he2016deep} design the Residual structure (ResBlock), which uses a skip-layer connection to strengthen the gradient signals between layers. However, while ResBlock has been widely used in GAN literature \citep{wang2018high,karras2020analyzing}, it also increases the computation cost. 

We reformulate the skip-connection idea with two unique designs into the \textit{Skip-Layer Excitation} module (SLE). First, ResBlock implements skip-connection as an element-wise addition between the activations from different conv-layers. It requires the spatial dimensions of the activations to be the same. Instead of addition, we apply channel-wise multiplications between the activations, eliminating the heavy computation of convolution (since one side of the activations now has a spatial dimension of $1^2$). Second, in prior GAN works, skip-connections are only used within the same resolution. In contrast, we perform skip-connection between resolutions with a much longer range (e.g., $8^2$ and $128^2$, $16^2$ and $256^2$), since an equal spatial-dimension is no longer required. The two designs make SLE inherits the advantages of ResBlock with a shortcut gradient flow, meanwhile without an extra computation burden.

Formally, we define the Skip-Layer Excitation module as:
\begin{align}
   \mathbf{y} = \mathcal{F}(\mathbf{x}_{low}, \{\mathbf{W}_{i}\}) \cdot \mathbf{x}_{high}
\end{align}
Here $\mathbf{x}$ and $\mathbf{y}$ are the input and output feature-maps of the SLE module, the function $\mathcal{F}$ contains the operations on $\mathbf{x}_{low}$, and $\mathbf{W}_i$ indicates the module weights to be learned. The left panel in Fig.~\ref{fig:g_overview} shows an SLE module in practice, where $\mathbf{x}_{low}$ and $\mathbf{x}_{high}$ are the feature-maps at $8\times8$ and $128\times128$ resolution respectively. An adaptive average-pooling layer in $\mathcal{F}$ first down-samples $\mathbf{x}_{low}$ into $4\times4$ along the spatial-dimensions, then a conv-layer further down-samples it into $1\times1$. A LeakyReLU is used to model the non-linearity, and another conv-layer projects $\mathbf{x}_{low}$ to have the same channel size as $\mathbf{x}_{high}$. Finally, after a gating operation via a Sigmoid function, the output from $\mathcal{F}$ multiplies $\mathbf{x}_{high}$ along the channel dimension, yielding $\mathbf{y}$ with the same shape as $\mathbf{x}_{high}$.

SLE partially resembles the Squeeze-and-Excitation module (SE) proposed by \citeauthor{hu2018squeeze}. However, SE operates within one feature-map as a self-gating module. In comparison, SLE performs between feature-maps that are far away from each other. While SLE brings the benefit of channel-wise feature re-calibration just like SE, it also strengthens the whole model's gradient flow like ResBlock. The channel-wise multiplication in SLE also coincides with Instance Normalization \citep{ulyanov2016instance,huang2017arbitrary}, which is widely used in style-transfer. Similarly, we show that SLE enables $G$ to automatically disentangle the content and style attributes, just like StyleGAN \citep{karras2019style}. As SLE performs on high-resolution feature-maps, altering these feature-maps is shown to be more likely to change the style attributes of the generated image \citep{karras2019style,liu2020time}. By replacing $\mathrm{x}_{low}$ in SLE from another synthesized sample, our $G$ can generate an image with the content unchanged, but in the same style of the new replacing image.

\subsection{Self-supervised discriminator}
Our approach to provide a strong regularization for $D$ is surprisingly simple. We treat $D$ as an encoder and train it with small decoders. Such auto-encoding training forces $D$ to extract image features that the decoders can give good reconstructions. The decoders are optimized together with $D$ on a simple reconstruction loss, which is only trained on real samples:
\begin{align}
    &\mathcal{L}_{recons}= \mathbb{E}_{ \mathbf{f} \sim D_{encode}(x), \; x \sim I_{real} } [ ||  \mathcal{G} ( \mathbf{f} ) - \mathcal{T}(x) || ],
    \label{eq:rec}
\end{align}

where $\mathbf{f}$ is the intermediate feature-maps from $D$, the function $\mathcal{G}$ contains the processing on $\mathbf{f}$ and the decoder, and the function $\mathcal{T}$ represents the processing on sample $x$ from real images $I_{real}$.

\begin{figure}[h]
\begin{center}
\includegraphics[width=0.9\linewidth,height=3.6cm]{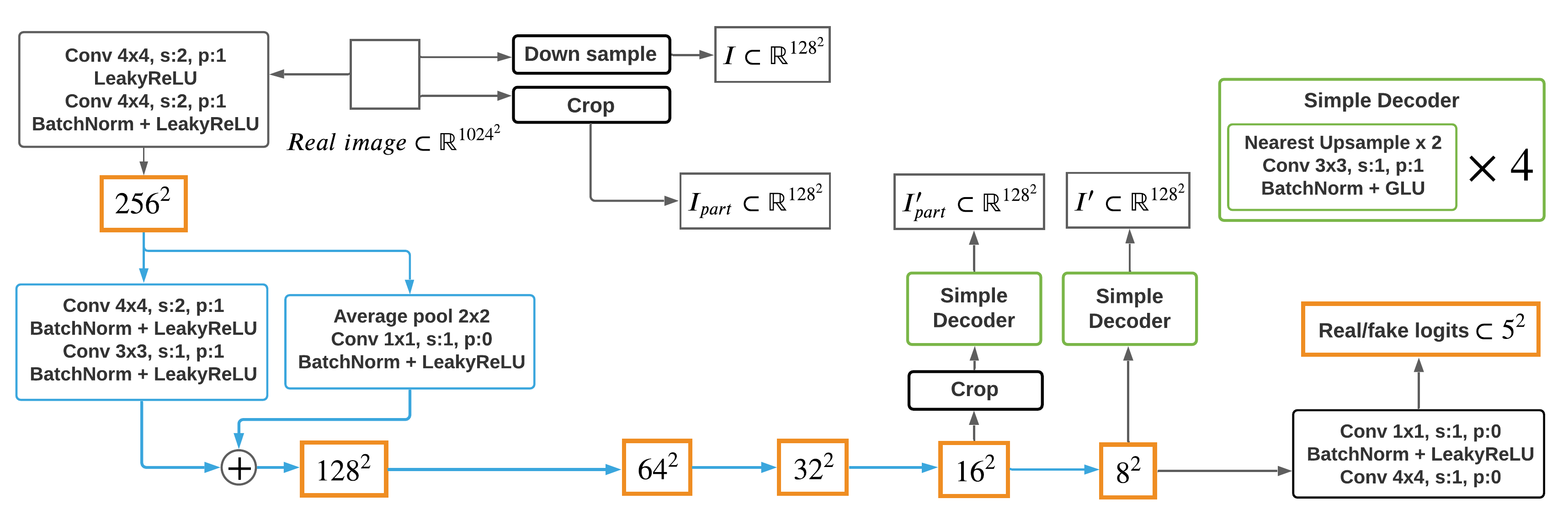}
\end{center}
\caption{The structure and the forward flow of the Discriminator. Blue box and arrows represent the same residual down-sampling structure, green boxes mean the same decoder structure.}
\label{fig:d_overview}
\end{figure}

Our self-supervised $D$ is illustrated in Fig.~\ref{fig:d_overview}, where we employ two decoders for the feature-maps on two scales: $\mathbf{f}_1$ on $16^2$ and $\mathbf{f}_2$ on $8^2$ . The decoders only have four conv-layers to produce images at $128\times128$ resolution, causing little extra computations (much less than other regularization methods). We randomly crop $\mathbf{f}_1$ with $\frac{1}{8}$ of its height and width, then crop the real image on the same portion to get $I_{part}$. We resize the real image to get $I$. The decoders produce $I'_{part}$ from the cropped $\mathbf{f}_1$, and $I'$ from $\mathbf{f}_2$. Finally, $D$ and the decoders are trained together to minimize the loss in eq.~\ref{eq:rec}, by matching $I'_{part}$ to $I_{part}$ and $I'$ to $I$.

Such reconstructive training makes sure that $D$ extracts a more comprehensive representation from the inputs, covering both the overall compositions (from $\mathbf{f}_2$) and detailed textures (from $\mathbf{f}_1$). Note that the processing in $\mathcal{G}$ and $\mathcal{T}$ are not limited to cropping; more operations remain to be explored for better performance. The auto-encoding approach we employ is a typical method for self-supervised learning, which has been well recognized to improve the model robustness and generalization ability \citep{he2020momentum,hendrycks2019using,jing2020self,goyal2019scaling}. In the context of GAN, we find that a regularized $D$ via self-supervision training strategies significantly improves the synthesis quality on $G$, among which auto-encoding brings the most performance boost.

Although our self-supervision strategy for $D$ comes in the form of an auto-encoder (AE), this approach is fundamentally different from works trying to combine GAN and AE \citep{larsen2016autoencoding,guo2019auto,zhao2016energy,berthelot2017began}. The latter works mostly train $ G $ as a decoder on a learned latent space from $ D $, or treat the adversarial training with $ D $ as an supplementary loss besides AE's training. In contrast, our model is a pure GAN with a much simpler training schema. The auto-encoding training is only for regularizing $D$, where $G$ is not involved.

In sum, we employ the hinge version of the adversarial loss (\cite{lim2017geometric,tran2017deep}) to iteratively train our D and G. We find the different GAN losses make little performance difference, while hinge loss computes the fastest: 
\begin{align}
    \mathcal{L}_{D}= &-  \mathbb{E}_{ x \sim I_{real} } [ min(0, -1 + D(x)) ]  -\mathbb{E}_{ \hat{x} \sim G(z) }[ min(0, -1 - D(\hat{x}) ]
    + \mathcal{L}_{recons} \\
    \mathcal{L}_{G}= &-\mathbb{E}_{ z \sim \mathcal{N} }[ D( G(z) ) ] ,
\end{align}

\section{Experiment}

\textbf{Datasets}: We conduct experiments on multiple datasets with a wide range of content categories. On $256\times256$ resolution, we test on Animal-Face Dog and Cat \citep{si2011learning}, 100-Shot-Obama, Panda, and Grumpy-cat \citep{zhao2020differentiable}. On $1024\times1024$ resolution, we test on Flickr-Face-HQ (FFHQ) \citep{karras2019style}, Oxford-flowers \citep{Nilsback06}, art paintings from WikiArt (wikiart.org), photographs on natural landscape from Unsplash (unsplash.com), Pokemon (pokemon.com), anime face, skull, and shell. These datasets are designed to cover images with different characteristics: photo realistic, graphic-illustration, and art-like images.

\textbf{Metrics}: We use two metrics to measure the models' synthesis performance: 1) Fréchet Inception Distance (FID)~\citep{heusel2017gans} measures the overall semantic realism of the synthesized images. For datasets with less than 1000 images (most only have 100 images), we let $G$ generate 5000 images and compute FID between the synthesized images and the whole training set. 2) Learned perceptual similarity (LPIPS)~\citep{zhang2018unreasonable} provides a perceptual distance between two images. We use LPIPS to report the reconstruction quality when we perform latent space back-tracking on $G$ given real images, and measure the auto-encoding performance. We find it unnecessary to involve other metrics, as FID is unlikely to be inconsistent with the others, given the notable performance gap between our model and the compared ones. For all the testings, we train the models 5 times with random seeds, and report the highest scores. The relative error is less than five percent on average.   

\textbf{Compared Models}: We compare our model with: 1) the state-of-the-art (SOTA) unconditional model, StyleGAN2, 2) a baseline model ablated from our proposed one. Note that we adopt StyleGAN2 with recent studies from \citep{karras2020training,zhao2020differentiable}, including the model configuration and differentiable data-augmentation, for the best training on few-sample datasets. Since StyleGAN2 requires much more computing-cost (cc) to train, we derive an extra baseline model. In sum, we compare our model with StyleGAN2 on the absolute image synthesis quality regardless of cc, and use the baseline model for the reference within a comparable cc range.

The baseline model is the strongest performer that we integrated from various GAN techniques based on DCGAN \citep{radford2015unsupervised}: 1) spectral-normalization \citep{miyato2018spectral}, 2) exponential-moving-average \citep{yazici2018unusual} optimization on $G$, 3) differentiable-augmentation, 4) GLU \citep{dauphin2017language} instead of ReLU in $G$. We build our model upon the baseline with the two proposed techniques: the skip-layer excitation module and the self-supervised discriminator. 

\begin{table}[h]
\caption{Computational cost comparison of the models.}
\label{table:compute-cost}
\begin{center}
 \resizebox{\linewidth}{!}{
\begin{tabular}{l|l| r r r r r}
\toprule
\multicolumn{2}{c}{}    & StyleGAN2@0.25 & StyleGAN2@0.5 & StyleGAN2 & Baseline & Ours \\
\cmidrule{1-7}
\multirow{3}{*}{\begin{tabular}[c]{@{}l@{}}Resolution: $256^2$ \\ Batch-size: 8\end{tabular}} & Training time  (hour / 10k iter) & 1                & 1.8             & 3.8           & 0.7      & 1    \\
& Training vram (GB)               & 7                & 16              & 18            & 5        & 6.5  \\ 
& Model parameters (million) & 27.557 & 45.029 & 108.843 & 44.359 & 47.363 \\
\midrule
\multirow{3}{*}{\begin{tabular}[c]{@{}l@{}}Resolution: $1024^2$\\ Batch-size: 8\end{tabular}} & Training time  (hour / 10k iter) & 3.6              & 5               & 7             & 1.3      & 1.7  \\
& Training vram (GB)               & 12               & 23              & 36            & 9        & 10   \\ 
& Model parameters (million) & 27.591 & 45.15 & 109.229 & 44.377 & 47.413 \\
\bottomrule
\end{tabular}
}
\end{center}
\end{table}

Table.~\ref{table:compute-cost} presents the normalized cc figures of the models on Nvidia's RTX 2080-Ti GPU, implemented using PyTorch \citep{paszke2017automatic}. Importantly, the slimed StyleGAN2 with $\frac{1}{4}$ parameters cannot converge on the tested datasets at $1024^2$ resolution. We compare to the StyleGAN2 with $\frac{1}{2}$ parameters (if not specifically mentioned) in the following experiments.

\subsection{Image synthesis performance}
\textbf{Few-shot generation}: Collecting large-scale image datasets are expensive, or even impossible, for a certain character, a genre, or a topic. On those few-shot datasets, a data-efficient model becomes especially valuable for the image generation task. In Table.~\ref{table:fid-256} and Table.~\ref{table:fid-1024}, we show that our model not only achieves superior performance on the few-shot datasets, but also much more computational-efficient than the compared methods. We save the checkpoints every 10k iterations during training and report the best FID from the checkpoints (happens at least after 15 hours of training for StyleGAN2 on all datasets). Among the 12 datasets, our model performs the best on 10 of them.

Please note that, due to the VRAM requirement for StyleGAN2 when trained on $1024^2$ resolution, we have to train the models in Table.~\ref{table:fid-1024} on a RTX TITAN GPU. In practice, 2080-TI and TITAN share a similar performance, and our model runs the same time on both GPUs.

\begin{table}[h]
\vspace{-0mm}
\caption{FID comparison at $256^2$ resolution on few-sample datasets.}
\label{table:fid-256}
\vspace{-2mm}
\begin{center}
 \resizebox{\linewidth}{!}{
\begin{tabular}{l|l|l|r r r r r}
\toprule
 \multicolumn{3}{c}{}   & Animal Face - Dog & Animal Face - Cat & Obama & Panda & Grumpy-cat \\
 \midrule
 \multicolumn{3}{c|}{Image number}  & 389 & 160 & 100 & 100 & 100            \\
 \midrule
 \multirow{6}{*}{\begin{tabular}[c]{@{}l@{}}Training time on\\ one RTX 2080-Ti\end{tabular}} & \multirow{2}{*}{20 hour} & StyleGAN2  & 58.85 & 42.44 & 46.87 & 12.06 & 27.08 \\
                               &    & StyleGAN2 finetune & 61.03 & 46.07  & \textbf{35.75} & 14.5 & 29.34 \\
 \cmidrule{2-8}
  & \multirow{4}{*}{5 hour}  & Baseline & 108.19 & 150.3 & 62.74  & 15.4  & 42.13          \\
                    & & Baseline+Skip & 94.21 & 72.97  & 52.50  & 14.39 & 38.17          \\
                & & Baseline+decode & 56.25 & 36.74  & 44.34  & 10.12 & 29.38          \\
 & & Ours (B+Skip+decode) & \textbf{50.66}  & \textbf{35.11} & 41.05 & \textbf{10.03} & \textbf{26.65} \\
 \bottomrule
\end{tabular}}
\end{center}
\vspace{-2mm}
\end{table}

\textbf{Training from scratch vs. fine-tuning}: Fine-tuning from a pre-trained GAN \citep{mo2020freeze,noguchi2019image,wang2020minegan} has been the go-to method for the image generation task on datasets with few samples. However, its performance highly depends on the semantic consistency between the new dataset and the available pre-trained model. According to \citeauthor{zhao2020differentiable}, fine-tuning performs worse than training from scratch in most cases, when the content from the new dataset strays away from the original one. We confirm the limitation of current fine-tuning methods from Table.~\ref{table:fid-256} and Table.~\ref{table:fid-1024}, where we fine-tune StyleGAN2 trained on FFHQ use the Freeze-D method from \citeauthor{mo2020freeze}. Among all the tested datasets, only Obama and Skull favor the fine-tuning method, making sense since the two sets share the most similar contents to FFHQ.

\textbf{Module ablation study}: We experiment with the two proposed modules in Table.~\ref{table:fid-256}, where both SLE (skip) and decoding-on-$D$ (decode) can separately boost the model performance. It shows that the two modules are orthogonal to each other in improving the model performance, and the self-supervised $D$ makes the biggest contribution. Importantly, the baseline model and StyleGAN2 diverge fast after the listed training time. In contrast, our model is less likely to mode collapse among the tested datasets. Unlike the baseline model which usually model-collapse after trained for 10 hours, our model maintains a good synthesis quality and won't collapse even after trained for 20 hours. We argue that it is the decoding regularization on $ D $ that prevents the model from divergence.

\begin{table}[h]
\vspace{-0mm}
\caption{FID comparison at $1024^2$ resolution on few-sample datasets.}
\vspace{-2mm}
\label{table:fid-1024}
\begin{center}
 \resizebox{\linewidth}{!}{
\begin{tabular}{l|l|l|r r r r r r r }
\toprule
 \multicolumn{3}{c}{}              & Art Paintings & FFHQ & Flower & Pokemon  & Anime Face & Skull & Shell  \\
 \midrule
 \multicolumn{3}{c|}{Image number} & 1000          & 1000 & 1000   & 800      & 120        & 100   & 60     \\
 \midrule
 \multirow{4}{*}{\begin{tabular}[c]{@{}l@{}}Training \\time on 
                                            one \\ RTX TITAN\end{tabular}}
                                   & \multirow{2}{*}{24 hour} 
                                                   & StyleGAN2 & 74.56 & 25.66 & 45.23 & 190.23 & 152.73 & 127.98 &241.37 \\
    &    & StyleGAN2 finetune & N/A & N/A & 36.72 & 60.12	& 61.23& 	\textbf{107.68}	& 220.45 \\
 \cmidrule{2-10}
  & \multirow{2}{*}{8 hour}  & Baseline & 62.27 & 38.35 & 42.25 & 67.86  & 101.23 & 186.45 & 202.32          \\
 & & Ours & \textbf{45.08} &	\textbf{24.45} & \textbf{25.66} & \textbf{57.19} &	\textbf{59.38} &	130.05 &	\textbf{155.47} \\
 \bottomrule
\end{tabular}}
\end{center}
\vspace{-0mm}

\end{table}

\begin{table}[h]
\vspace{-2mm}
\caption{FID comparison at $1024^2$ resolution on datasets with more images.}
\label{table:fid-1024-more}
\vspace{-2mm}
\begin{center}
 \resizebox{0.96\linewidth}{!}{
\begin{tabular}{l | l | l l l | l l l l | l l l}
\toprule
 \multirow{2}{*}{\begin{tabular}[c]{@{}l@{}} Model \end{tabular}} & Dataset
 & \multicolumn{3}{c}{Art Paintings} & \multicolumn{4}{c}{FFHQ} & \multicolumn{3}{c}{Nature Photograph} \\
 \cmidrule{2-12}
 & Image number & 2k & 5k & 10k & 2k & 5k & 10k & 70k & 2k & 5k & 10k\\
 \midrule
 \multicolumn{2}{l|}{StyleGAN2} & 70.02          & 48.36         & \textbf{41.23} & \textbf{18.38} & \textbf{10.45} & \textbf{7.86} & \textbf{4.4} & 67.12 & \textbf{41.47} & \textbf{39.05} \\
 \midrule
 \multicolumn{2}{l|}{Baseline} & 60.02           & 51.23         & 49.38          & 36.45 & 27.86 & 25.12 & 17.62 & 71.47 & 66.05  & 62.28\\
 \midrule 
 \multicolumn{2}{l|}{Ours}     & \textbf{44.57} & \textbf{43.27} & 42.53          & 19.01	& 17.93 & 16.45 & 12.38 & \textbf{52.47} & 45.07 & 43.65  \\
 \bottomrule
\end{tabular}}
\end{center}
\vspace{-0mm}

\end{table}

\textbf{Training with more images}: For more thorough evaluation, we also test our model on datasets with more sufficient training samples, as shown in Table.~\ref{table:fid-1024-more}. We train the full StyleGAN2 for around five days on the Art and Photograph dataset with a batch-size of 16 on two TITAN RTX GPUs, and use the latest official figures on FFHQ from \citeauthor{zhao2020differentiable}. Instead, we train our model for only 24 hours, with a batch-size of 8 on a single 2080-Ti GPU. Specifically, for FFHQ with all 70000 images, we train our model with a larger batch-size of 32, to reflect an optimal performance of our model. 

In this test, we follow the common practice of computing FID by generating 50k images and use the whole training set as the reference distribution. Note that StyleGAN2 has more than double the parameters compared to our model, and trained with a much larger batch-size on FFHQ. These factors contribute to its better performances when given enough training samples and computing power. Meanwhile, our model keeps up well with StyleGAN2 across all testings with a considerably lower computing budget, showing a compelling performance even on larger-scale datasets, and a consistent performance boost over the baseline model.

\begin{figure}
\begin{center}
\includegraphics[width=1\linewidth,height=1.46\linewidth]{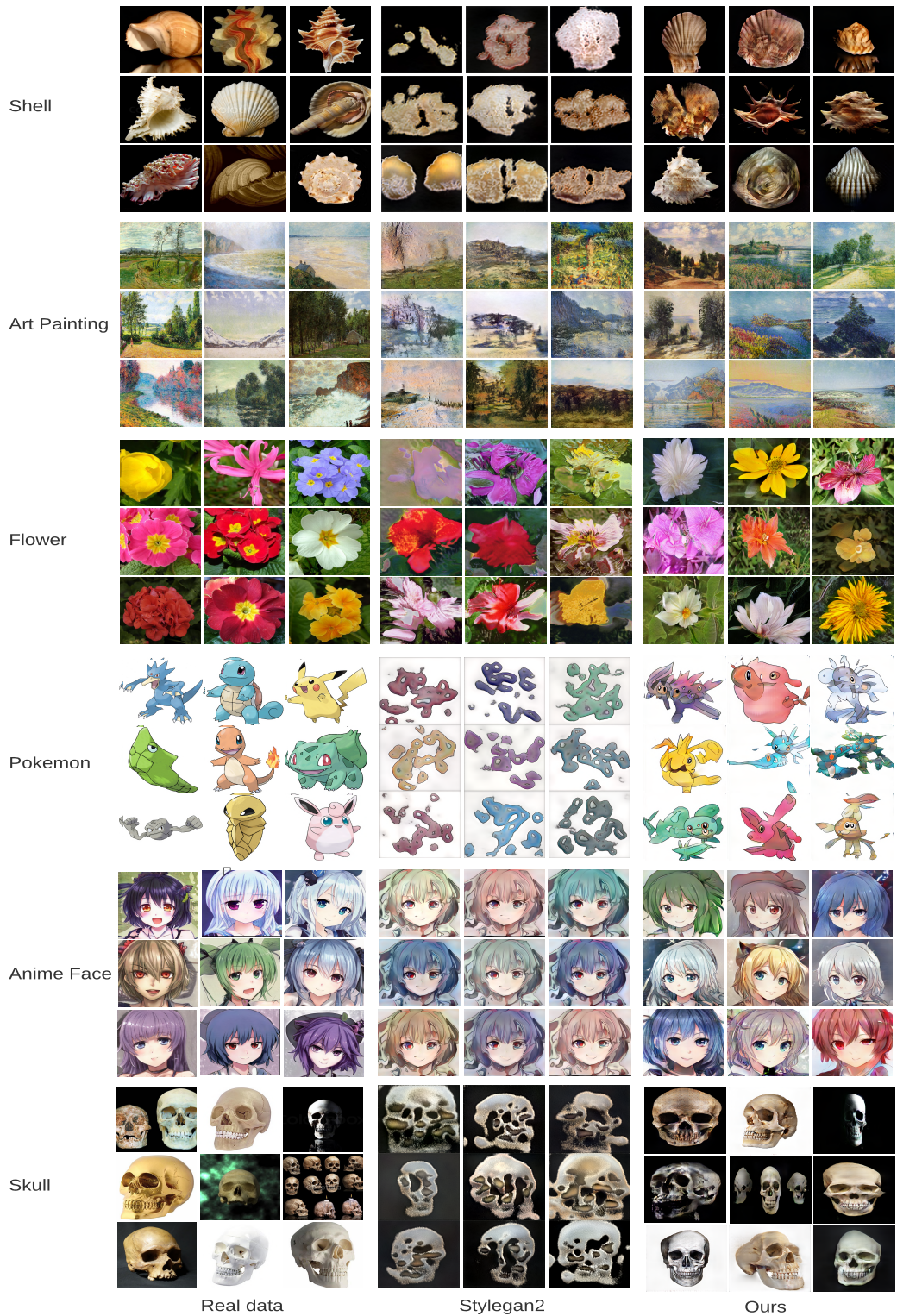}
\end{center}
\caption{\textbf{Qualitative comparison between our model and StyleGAN2} on $1024^2$ resolution datasets. The left-most panel shows the training images, and the right two panels show the uncurated samples from StyleGAN2 and our model. Both models are trained from scratch for 10 hours with a batch-size of 8. The samples are generated from the checkpoint with the lowest FID. }
\label{fig:quality-compare}
\end{figure}

\textbf{Qualitative results}: The advantage of our model becomes more clear from the qualitative comparisons in Fig.~\ref{fig:quality-compare}. Given the same batch-size and training time, StyleGAN2 either converges slower or suffers from mode collapse. In contrast, our model consistently generates satisfactory images. Note that the best results from our model on Flower, Shell, and Pokemon only take three hours' training, and for the rest three datasets, the best performance is achieved at training for eight hours. For StyleGAN2 on “shell”, “anime face”, and “Pokemon”, the images shown in Fig.~\ref{fig:quality-compare} are already from the best epoch, which they match the scores in Table.~\ref{table:fid-256} and Table.~\ref{table:fid-1024}. For the rest of the datasets, the quality increase from StyleGAN2 is also limited given more training time. 

\subsection{More Analysis and Applications}

\begin{table}
	
	\begin{minipage}{0.56\linewidth}
		\centering
		\hspace{-1cm}
		\includegraphics[width=\linewidth,height=0.8\linewidth]{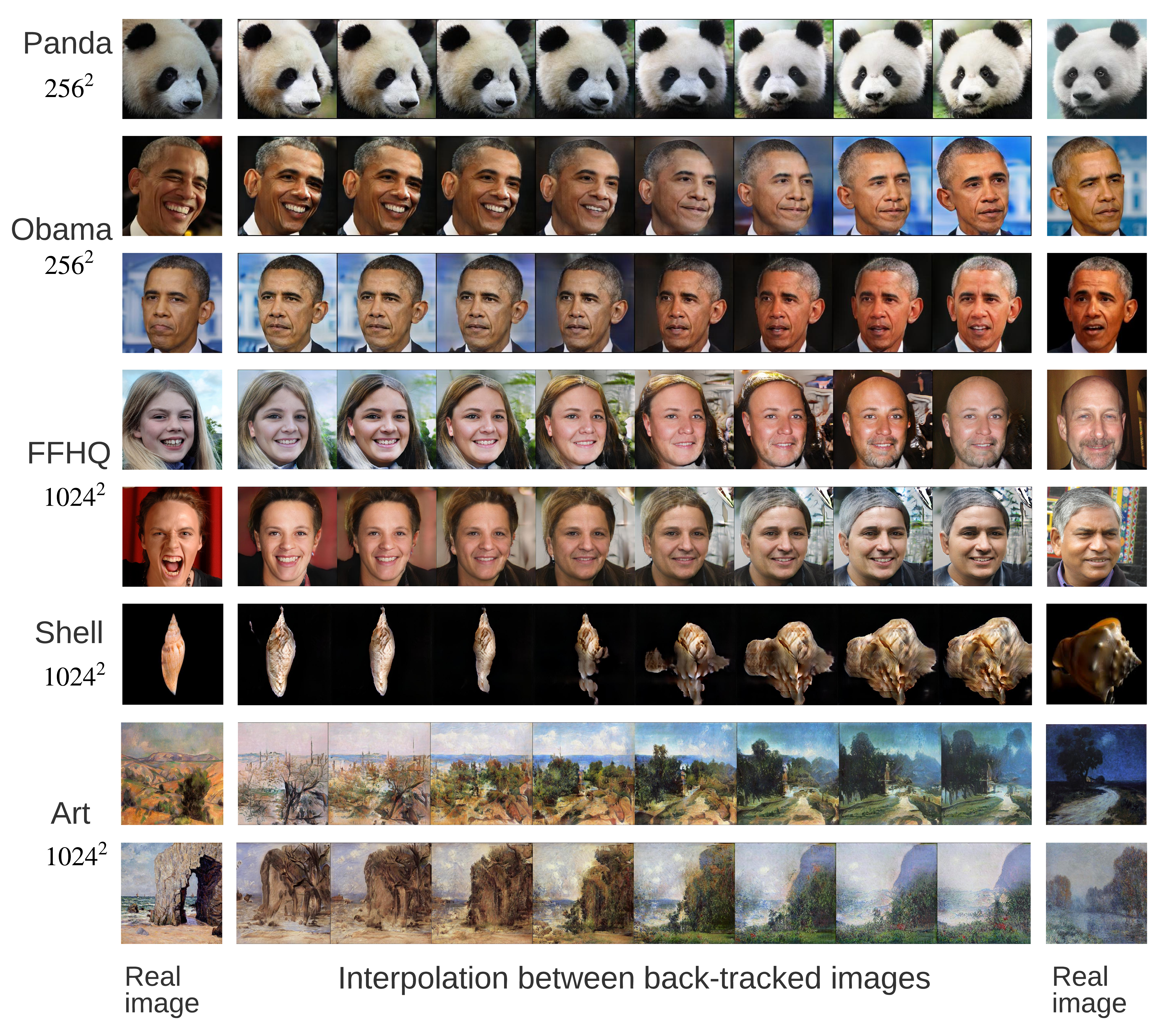}
		\captionof{figure}{Latent space back-tracking and interpolation.}
		\label{fig:back-track-interpolate}
		
	\end{minipage}\hfill
	\begin{minipage}{0.4\linewidth}
		\caption{LPIPS of back-tracking with $G$}
        \label{table:lpips-back-track}
        \resizebox{\linewidth}{!}{
        \begin{tabular}{l | l l l l}
        \toprule
        & Cat & Dog & FFHQ & Art \\
        \midrule
        Resolution & \multicolumn{2}{c|}{256} & 
        \multicolumn{2}{c}{1024} \\
        \midrule
        Baseline @ 20k iter & 2.113 & 2.073 & 2.589 & 2.916\\
        Baseline @ 40k iter & 2.513 & 2.171 & 2.583 & 2.812\\
        Ours @ 40k iter & \textbf{1.821} & \textbf{1.918} & 2.425 & 2.624\\
        Ours @ 80k iter & 1.897 & 1.986 & \textbf{2.342} & \textbf{2.601} \\
        \bottomrule
        \end{tabular}}
        
        \vspace{8mm}
        
        \caption{FID of self-supervisions for $D$}
        \label{table:fid-self-supervise}
        \resizebox{\linewidth}{!}{
            \begin{tabular}{l | c c }
        \toprule
        & Art paintings & Nature photos \\
        \midrule
        a. contrastive loss & 47.14 & 57.04\\
        b. predict aspect ratio & 49.21 & 59.22 \\
        c. auto-encoding & \textbf{42.53} & \textbf{43.65} \\
        d. a+b & 46.02 & 54.23 \\
        e. a+b+c & 44.21 & 47.65 \\
        \bottomrule
        \end{tabular}}
	\end{minipage}
\end{table}

\textbf{Testing mode collapse with back-tracking}: From a well trained GAN, one can take a real image and invert it back to a vector in the latent space of $ G $, thus editing the image's content by altering the back-tracked vector. Despite the various back-tracking methods \citep{zhu2016generative,lipton2017precise,zhu2020domain,abdal2019image2stylegan}, a well generalized $G$ is arguably as important for the good inversions. 
To this end, we show that our model, although trained on limited image samples, still gets a desirable performance on real image back-tracking.

In Table~\ref{table:lpips-back-track}, we split the images from each dataset with a training/testing ratio of 9:1, and train $G$ on the training set. We compute a reconstruction error between all the images from the testing set and their inversions from $G$, after the same update of 1000 iterations on the latent vectors (to prevent the vectors from being far off the normal distribution). The baseline model's performance is getting worse with more training iterations, which reflects mode-collapse on $ G $. In contrast, our model gives better reconstructions with consistent performance over more training iterations. Fig.~\ref{fig:back-track-interpolate} presents the back-tracked examples (left-most and right-most samples in the middle panel) given the real images. The smooth interpolations from the back-tracked latent vectors also suggest little mode-collapse of our $G$ \citep{radford2015unsupervised,zhao2020differentiable,robb2020few}.

In addition, we show qualitative comparisons in appendix D, where our model maintains a good generation while StyleGAN2 and baseline are model-collapsed.

\textbf{The self-supervision methods and generalization ability  on $D$}: Apart from the auto-encoding training for $ D $, we show that $ D $ with other common self-supervising strategies also boost GAN's performance in our training settings. We test five self-supervision settings, as shown in Table~\ref{table:fid-self-supervise}, which all brings a substantial performance boost compared to the baseline model. Specifically, setting-a refers to contrastive learning which we treat each real image as a unique class and let $ D $ classify them. For setting-b, we train $D$ to predict the real image's original aspect-ratio since they are reshaped to square when fed to $D$. Setting-c is the method we employ in our model, which trains $ D $ as an encoder with a decoder to reconstruct real images. To better validate the benefit of self-supervision on $D$, all the testings are conducted on full training sets with 10000 images, with a batch-size of 8 to be consistent with Table~\ref{table:fid-1024-more}. We also tried training with a larger batch-size of 16, which the results are consistent to the batch-size of 8.

\begin{figure}
\centering
\includegraphics[width=1\linewidth,height=0.6\linewidth]{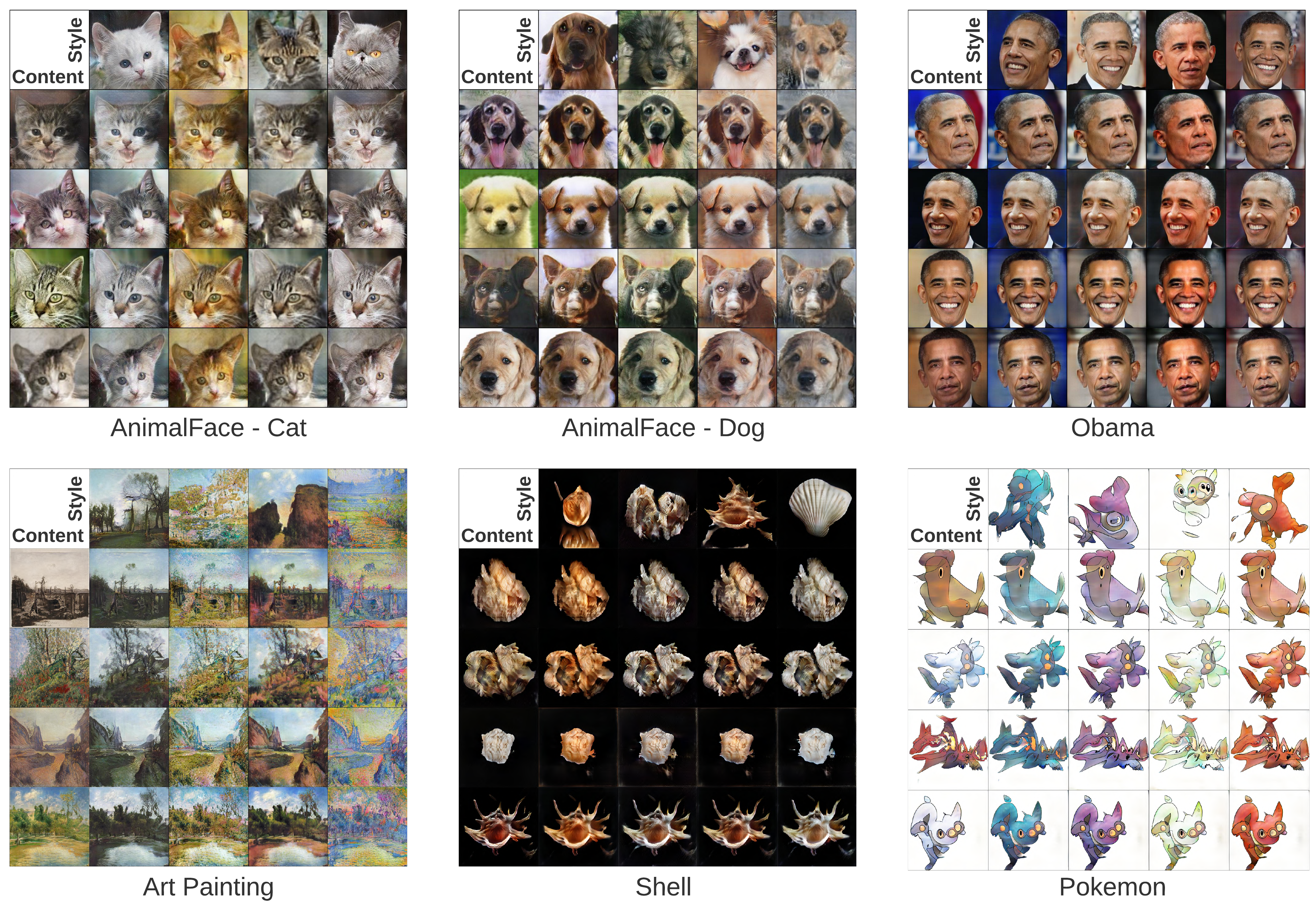}
\caption{\textbf{Style-mixing results} from our model trained for only 5 hours on single GPU. }
\label{fig:style-mixing}
\end{figure}

Interestingly, according to Table~\ref{table:fid-self-supervise}, while setting-c performs the best, combining it with the rest two settings lead to a clear performance downgrade. The similar behavior can be found on some other self-supervision settings, e.g. when follow \cite{chen2019self} with a "rotation-predicting" task on art-paintings and FFHQ datasets, we observe a performance downgrade even compared to the baseline model. We hypothesis the reason being that the auto-encoding forces $D$ to pay attention to more areas of the input image, thus extracts a more comprehensive feature-map to describe the input image (for a good reconstruction). In contrast, a classification task does not guarantee $D$ to cover the whole image. Instead, the task drives $D$ to only focus on small regions because the model can find class cues from small regions of the images. Focusing on limited regions (i.e., react to limited image patterns) is a typical overfitting behavior, which is also widely happening for $D$ in vanilla GANs. More discussion can be found in appendix B.

\textbf{Style mixing like StyleGAN}. With the channel-wise excitation module, our model gets the same functionality as StyleGAN: it learns to disentangle the images' high-level semantic attributes (style and content) in an unsupervised way, from $G$'s conv-layers at different scales. The style-mixing results are displayed in Fig.~\ref{fig:style-mixing}, where the top three datasets are $256\times256$ resolution, and the bottom three are $1024\times1024$ resolution. While StyleGAN2 suffers from converging on the bottom high-resolution datasets, our model successfully learns the style representations along the channel dimension on the ``excited" layers (i.e., for feature-maps on $256\times256$, $512\times512$ resolution). Please refer to appendix A and C for more information on SLE and style-mixing.

\section{Conclusion}
We introduce two techniques that stabilize the GAN training with an improved synthesis quality, given sub-hundred high-fidelity images and a limited computing resource. On thirteen datasets with a diverse content variation, we show that a skip-layer channel-wise excitation mechanism (SLE) and a self-supervised regularization on the discriminator significantly boost the synthesis performance of GAN. Both proposed techniques require minor changes to a vanilla GAN, enhancing GAN's practicality with a desirable plug-and-play property. We hope this work can benefit downstream tasks of GAN \citep{liu2020oogan,Liu_2020_ACCV,elgammal2017can} and provide new study perspectives for future research.

\bibliography{refs}
\bibliographystyle{iclr2021_conference}

\clearpage

\appendix

\noindent\textbf{Appendix}

\section{Performance boost from skip-layer excitation}

\begin{figure}[h]
\centering
\includegraphics[width=0.86\linewidth,height=0.46\linewidth]{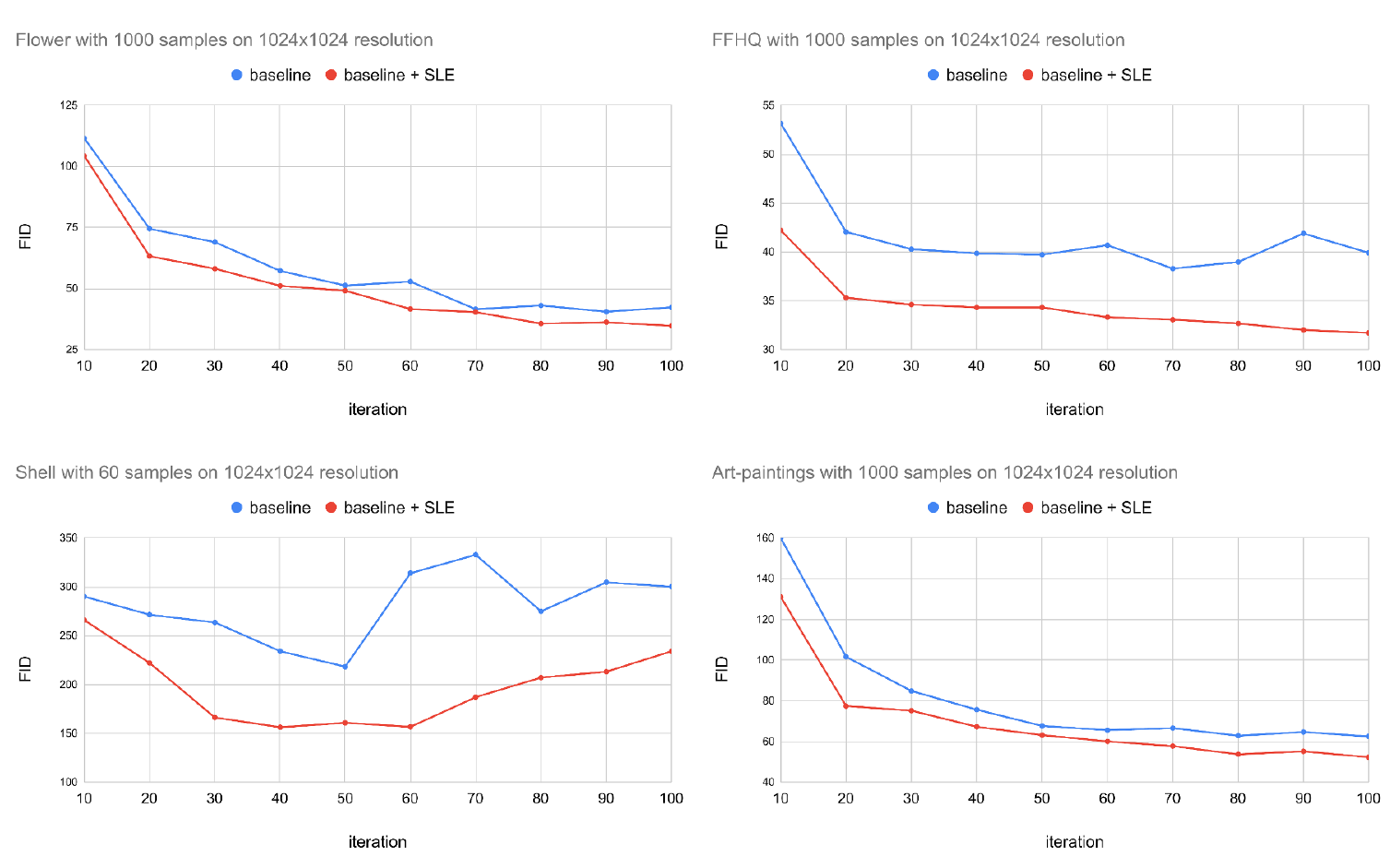}
\caption{\textbf{Ablation study for SLE module} on $1024\times1024$ resolution datasets. Each unit on the x-axis represents 1000 training iterations, and y-axis represents the FID score. }
\label{fig:sle-curve}
\end{figure}

Here we present a more detailed ablation study for the skip-layer excitation (SLE) module. We compare between the baseline model and the baseline equipped with SLE. On four $1024\times1024$ resolutions datasets: Flower, FFHQ, Shell and Art-paintings, we record the FID performance every 10000 iterations for every model. As shown in Fig.~\ref{fig:sle-curve}, SLE brings a constant performance boost on the baseline model over all iterations.

Our key observation is, SLE speeds up the convergence of GAN, where the most noticeable effect happens at the beginning of the training. In the first 20000 iterations, the generator $G$ is able to converge faster and reach to a good point where the baseline model needs much more training iterations to reach. On the other hand, although SLE provides a faster convergence on $G$, the overall model behavior with SLE seems follow the baseline model quite well, with a slightly better overall performance. 

In other words, the lines for the two models are parallel in each sub-plot in Fig.~\ref{fig:sle-curve}. Specifically, on Shell, the model with SLE also collapsed after 60000 iterations training, just like the baseline model. And on the rest three datasets, the FID improves much slower and almost stop changing in the later half training iterations. We think such model behavior makes sense, because the SLE module neither increases the model capacity (have very few parameter increase) nor exert any explicit regularization or guidance on the training of GAN. Therefore, SLE is unlikely to make a big difference after the model reaches a good converged state.

On the other hand, SLE does a good job speeding up the convergence for $G$, and improves the performance of $G$. More importantly, it is SLE that enables the unsupervised style-content disentanglement for our model, in a simpler and more cost-efficient way than StyleGAN and StyleGAN2.

\section{Feature-extraction performance of Discriminator}
\begin{table}[h]
\caption{LPIPS on $D$'s feature-extracting performance}
\label{table:d-decoder}
\begin{center}
\resizebox{0.9\linewidth}{!}{
\begin{tabular}{l | l l |c c c  | c c }
\toprule
& Grumpy Cat & Obama & \multicolumn{3}{c}{FFHQ} & \multicolumn{2}{c}{Art} \\
\midrule
Image number & 100 & 100 & 
1k & 70k & 0 & 1k & 0\\
\midrule
StyleGAN2 & 0.914 & 0.652 & 3.177 & 2.43 & 2.289  & 3.051 & 2.761 \\
\midrule
Baseline & 1.632 & 0.733 & 2.421 & N/A & 1.943 & 2.677 & 2.421\\
Baseline + Contrastive & 1.251 & 0.647 & 1.821 & N/A & 1.943 & 2.124 & 2.421\\
Baseline + AE & \textbf{0.725} & \textbf{0.405} & \textbf{1.075} & N/A & 1.943 & \textbf{1.806} & 2.421\\
Baseline + AE + Contrastive & 1.156 & 0.578 & 1.345 & N/A & 1.943 & 1.927 & 2.421\\
\bottomrule
\end{tabular}
}
\end{center}
\end{table}

Here we continue the discussion on the effectiveness of the self-supervised auto-encoding training for the discriminator $D$. Specifically, we explore the relationship between the \textit{ feature-extracting behavior on the discriminator $D$ } and the \textit{ synthesis performance of GAN }. By feature-extracting performance, we mean how comprehensive the feature-maps extracted by $D$ cover the information from the input images. This feature-extracting performance can be easily checked via an auto-encoding training. In detail, we take $D$ trained in GAN and fix it, then train a decoder for $D$ which tries to reconstruct the images from the feature-maps encoded by $D$. The intuition is, if $D$ pays attention to all the regions of an input image, and encode the image with a minimum information lost, then the decoder is easier to reconstruct the images encoded by $D$. In contrast, if $D$ is overfitting and only focus on limited local patterns of the images, the it outputs feature-maps with lost information, thus a decoder is unable to reconstruct the images from $D$'s output feature-map.  

We extract the second-last layer's activation for the decoder, which is the one for $D$ to determine the real/fake of an image. Table~\ref{table:d-decoder} shows the result, where we train all the decoders for the same 100000 iterations (all the decoders are converged). Note that such feature-extracting performance on $D$ does not necessarily imply a better synthesis performance for $G$. Moreover, the $D$ from StyleGAN2 is not comparable to the $D$ from baseline, since they have totally different model structure and complexity. 

Instead, according to Table~\ref{table:d-decoder}, we can get some interesting information. Firstly, the GAN training is actually making $D$ performs worse as a feature-encoder. According to row. 3 (StyleGAN2) and row. 4
(baseline), we find that the $D$ after a GAN training extracts less meaningful features compared to a randomly initialized $D$ (col. 6 and col. 8). It means that while the GAN training leads $D$ to find the discriminative features between the real and fake samples, it also effective let $D$ to ignore quite amount of information from the input images.

Secondly, we compare the baseline model to the ones with self-supervised learning guidance (row. 4,5,6,7). It shows that the self-supervisions on $D$ indeed lead to a more descriptive feature-extraction compared to the randomly initialization on $D$. Moreover, contrastive learning may also result in overfitting, since only a partial image (some local patterns) may be enough for the classification task. In comparison, the reconstruction task is more likely to let $D$ cover more information from the input images. To our surprise, combining auto-encoding training and the contrastive learning result in a worse performance on $D$. It shows that the classification objective affects the auto-encoding objective and changes the behavior of $D$, in a negative way. 

Last but not least, we do find that a better \textit{ feature-extracting performance on D} result in a better \textit{ synthesis performance of GAN }. And it seems true for both StyleGAN2 and the baseline model. For StyleGAN2 trained on FFHQ, $D$ trained with more data indeed preserves more information from the input images than $D$ trained on only 1000 images. For our baseline model, the feature-extracting performance on D aligns well with the respective FID scores. Besides, the self-supervision methods all effectively letting $D$ extracts more information from the images, compared to the randomly initialization and the vanilla GAN training.

Apart from the observations, we would like to emphasize that the experiments are mostly conducted on few-shot datasets. The results does not give a full picture of the relationship between the feature-extraction performance on $D$ and the synthesis performance of GAN, further study on larger-scale datasets are required. However, the experiments do validate the effectiveness of the self-supervision strategies on $D$ for an enhanced performance of GAN, on few-shot datasets.

\section{Style-mixing on different resolutions}

\begin{figure}[h]
\vspace{-2mm}
\centering
\includegraphics[width=\linewidth]{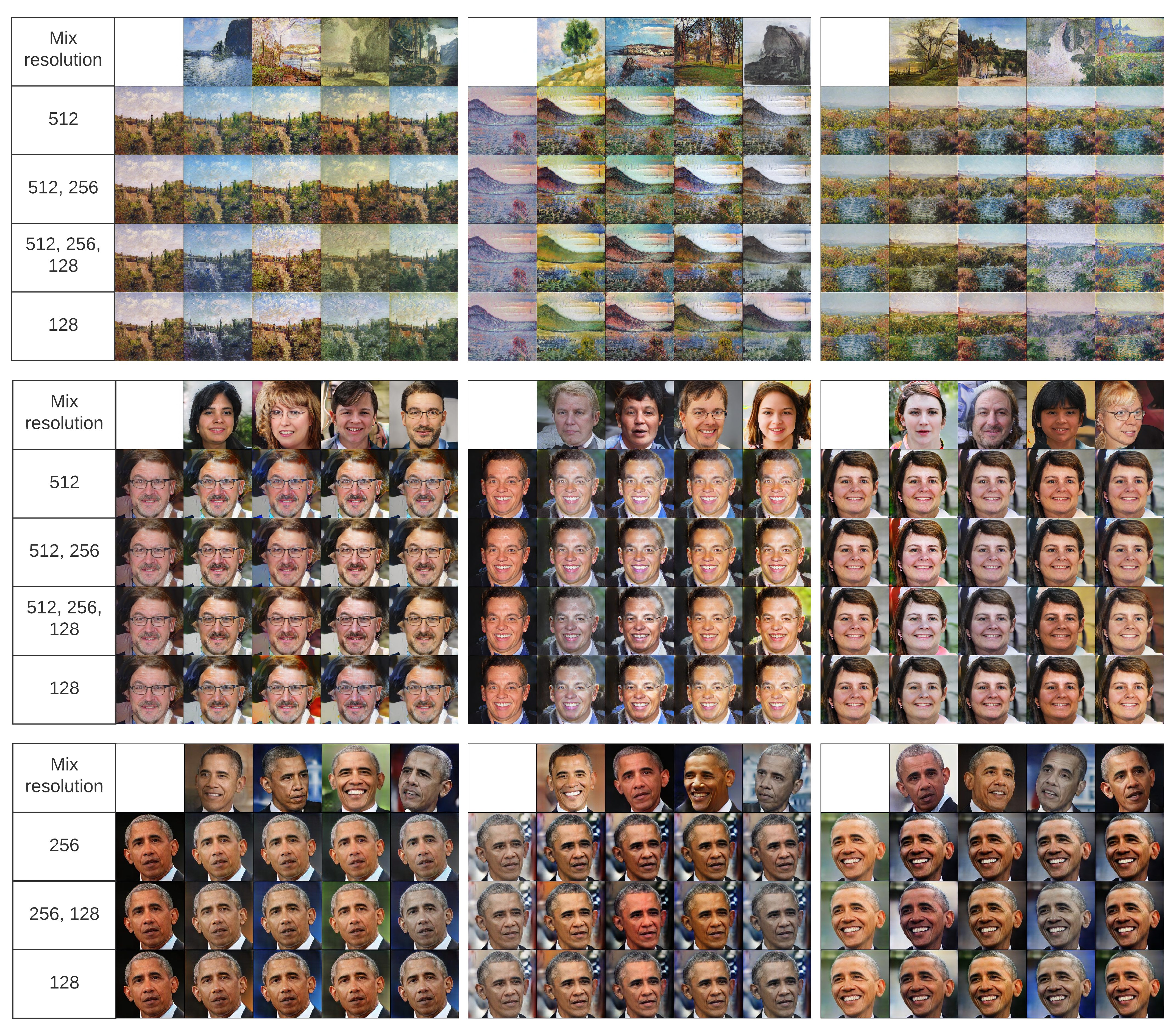}
\caption{\textbf{Style-mixing results} by swapping the features for SLE on different resolutions.}
\label{fig:style-mixing-multilayer}
\end{figure}

Here we present more qualitative results on the style-mix performance of our model. For the model trained on $1024\times1024$ resolution, there are three SLE layers that we can swap the feature-maps between generated samples, and there are two SLE layers for model on $256\times256$ resolution. 

Fig.~\ref{fig:style-mixing-multilayer} shows the results from the 1000 samples training on Art paintings and FFHQ at $1024\times1024$ resolution, and the 100 samples Obama at $256\times256$ resolution. In each row, we swap the $\mathbf{x}_{low}$ in the SLE layer from the image in col. 1 to the one from each image on row. 1. The best style-mixing results is achieved when the feature-map swapping is done on all resolutions. And the most effective layer that causes the most style changes is the layer on $128$ resolution. On $256\times256$ resolution, the model behaviors the same, where the SLE on lower resolution makes the most style difference.

On the Art-paintings data, the model performs well on style-mixing, where not only the coloring but also the texture can be controlled. The models transfers the style of flat or pointy brush stroke among the style-mixed synthetic images. However, the model does not perform as well on the FFHQ data. There are some cases where even the hair color can not be properly transferred. We speculate that the worse performance on FFHQ is due to the limited training sample and the dramatically varied background. The is no clear relationship between the front-end face and the background contents given the limited training samples, which confuses the model to disentangle more detailed style attributes. In contrast, Art-paintings have consistent style cues within each image and obvious connections between each object inside a scene, making it arguably easier than the FFHQ data. On the other hand, the model performs great on Obama given a even less 100 training images. It successfully transfers the style for both the face attributes and the background. Learning on $256^2$ resolution is a simpler task, and the model capacity is more sufficient on only 100 samples.  

\clearpage
\section{More Qualitative Comparison}

\begin{figure}[h]
\vspace{-2mm}
\centering
\includegraphics[width=\linewidth,height=1.3\linewidth]{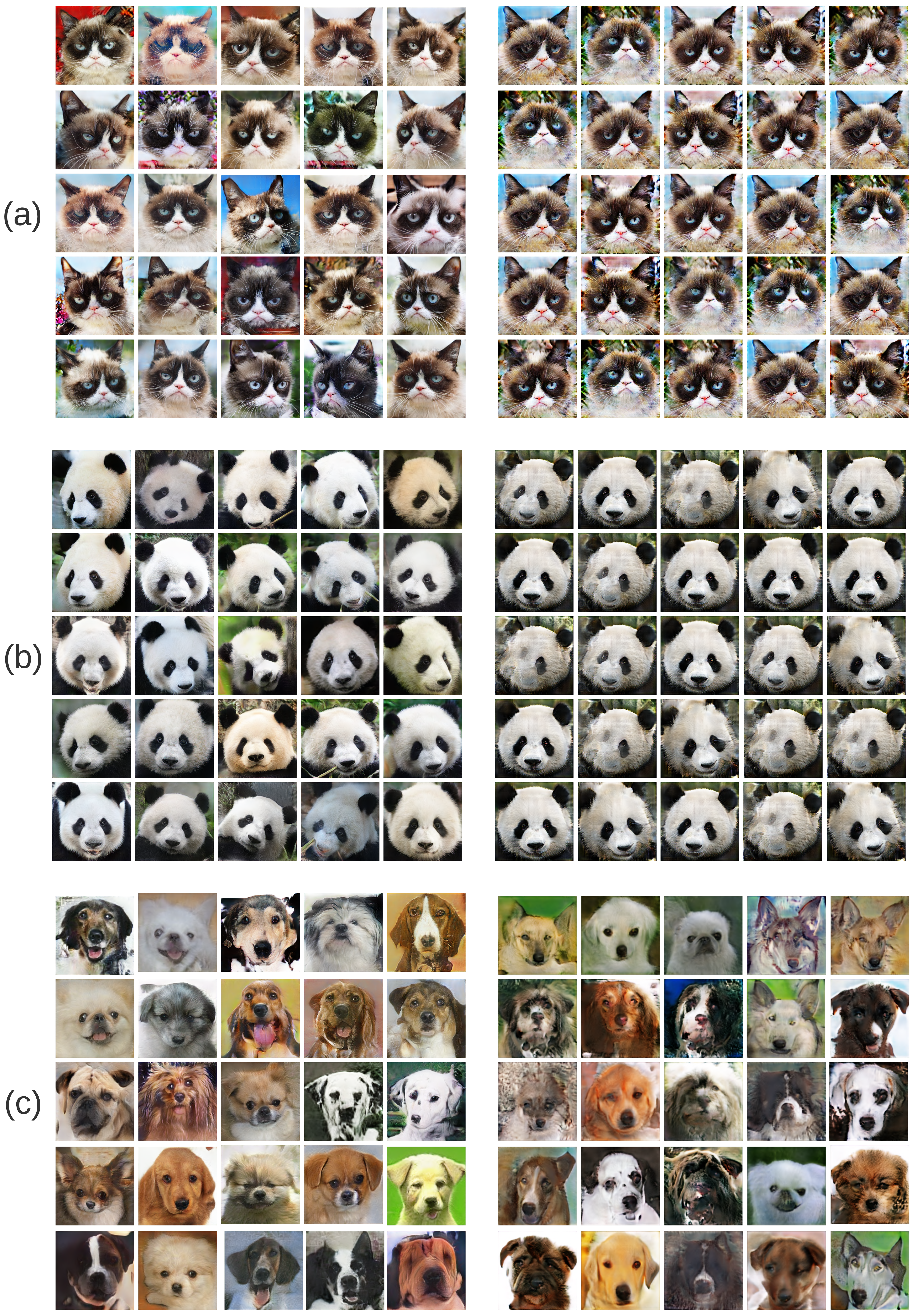}
\caption{\textbf{Comparison between our model and the baseline} For each dataset, the images are generated by the same set of randomly sampled noises. Images from our model is shown on the left, and the baseline results are on the right. All the model are trained for 50000 iterations with batch size of 8, which is more than enough for both models to converge. On (a) Grumpy-cat and (b) Panda, baseline model shows a clear mode collapse, while our model is generating diverse images; on (c) Animalface-dog, although not mode collapse, the baseline model shows a clear quality disadvantage compared to our model.}
\label{fig:style-mixing-multilayer}
\end{figure}

\begin{figure}[h]
\vspace{-2mm}
\centering
\includegraphics[width=\linewidth]{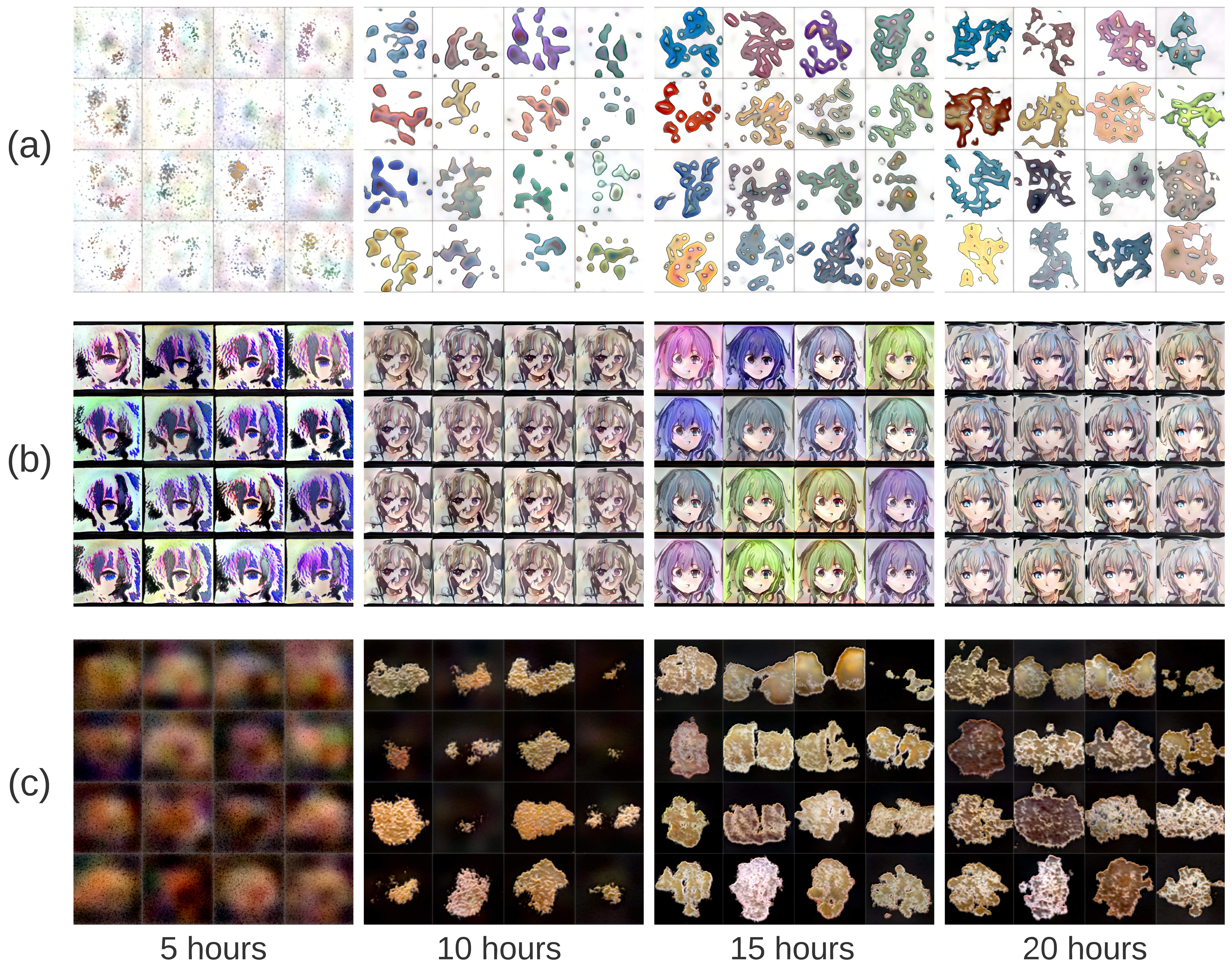}
\caption{\textbf{StyleGAN2 results during training} We show the results of the slimed StyleGAN2 at half channel numbers. StyleGAN2 converges much slower than our model on dataset (a) Pokemon and (c) Shell, and mode collapsed on (b) Anime-Face. }
\label{fig:style-mixing-multilayer}
\end{figure}

\clearpage

\section{Nearest images from training sets}

\begin{figure}[h]
\vspace{-2mm}
\centering
\includegraphics[width=\linewidth,height=1.3\linewidth]{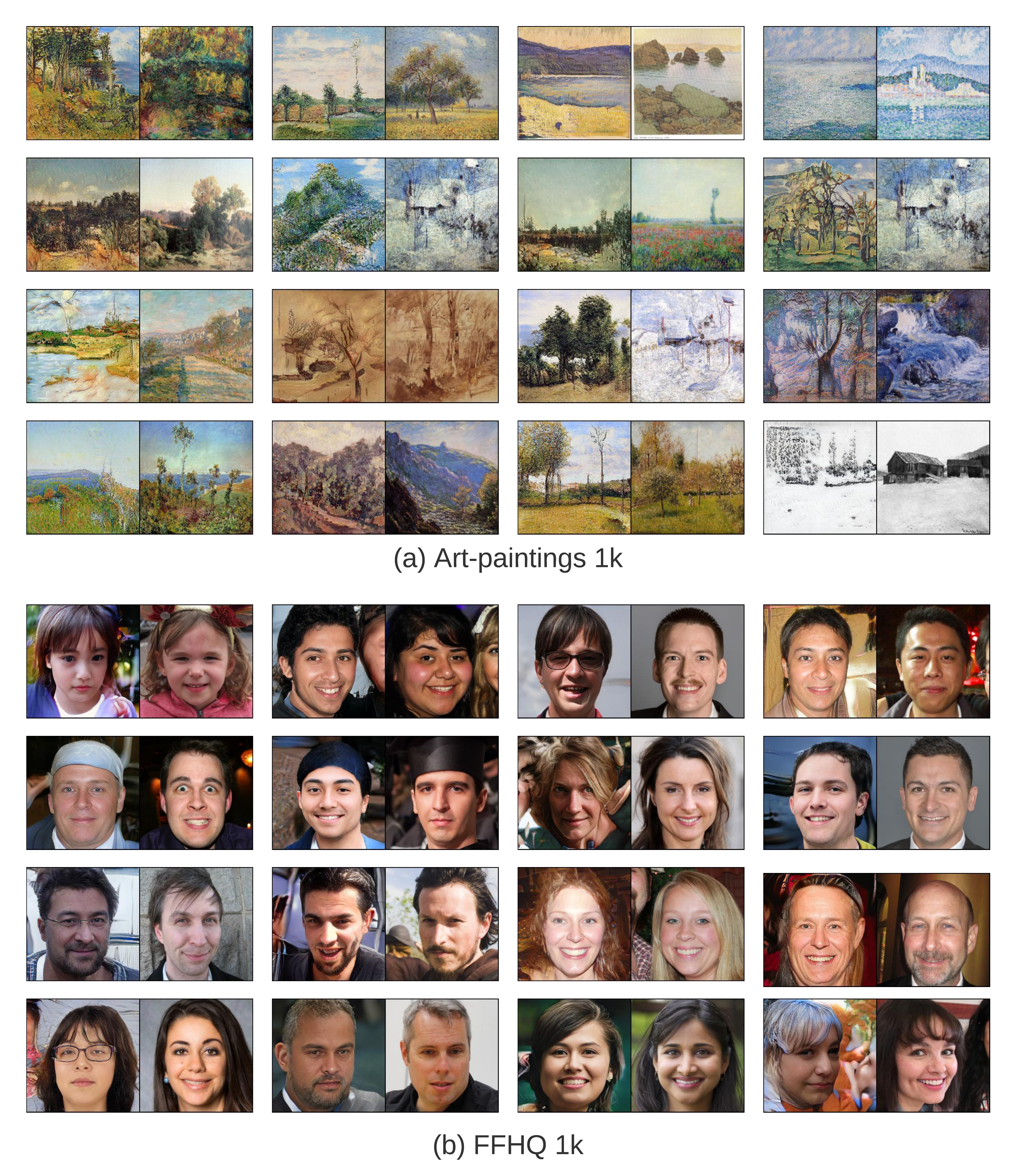}
\caption{\textbf{Nearest real images to the synthesized ones trained on 1000 images} For each pair of images, the left is the synthesized image from our model, and the right image is the closest image found from the real training data ranked by LPIPS score. The samples are uncurated, and our model is able to create new contents that well fitted to the training domain.}
\label{fig:nr_1}
\end{figure}
\clearpage

\begin{figure}[h]
\vspace{-2mm}
\centering
\includegraphics[width=\linewidth,height=1.2\linewidth]{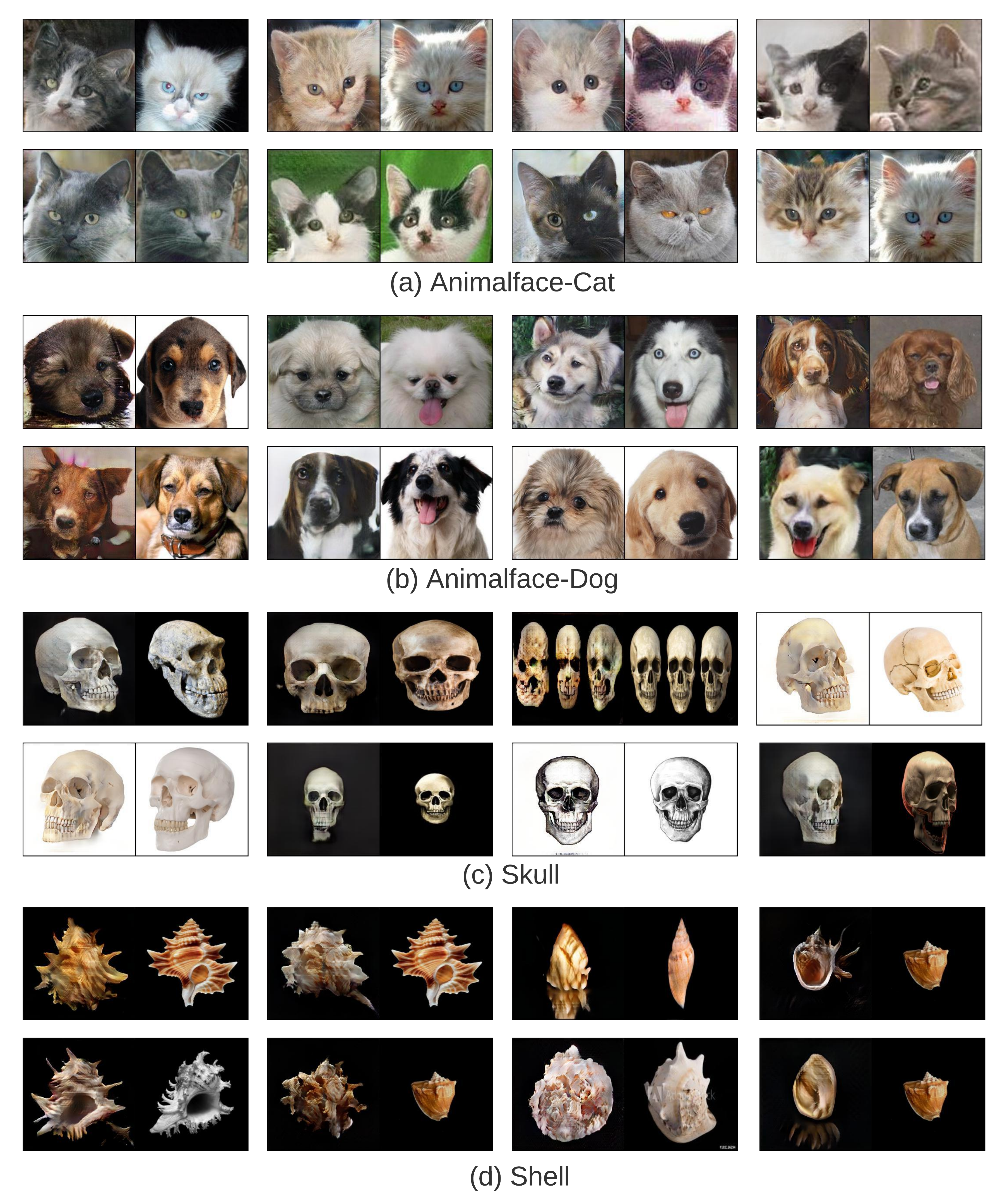}
\caption{\textbf{Nearest real images to the synthesized ones trained on 100 images} For each image pair, the left is the synthesized image from our model, and the right is the closest image found from the real training data ranked by LPIPS score. Even with only 100 training samples, these uncurated samples show our model is still able to combine the features learned from the real samples and synthesize new compositions.}
\label{fig:nr_1}
\end{figure}
\clearpage

\begin{table}[h]
\caption{LPIPS between synthetic images and their closest real images.}
\label{table:nn_lpips}
\begin{center}
 \resizebox{\linewidth}{!}{
\begin{tabular}{l | l l l l l l }
\toprule
 & Art paintings 1k	& FFHQ 1k & Skull &	Cat	& Dog & Shell \\
 \midrule
 augmented & 0.5499                                                    & 0.5279                                           & 0.389                                          & 0.3898                                       & 0.3847                                       & 0.3853                                         \\
 Our G         & 0.637                                                     & 0.5859                                           & 0.3168                                         & 0.5486                                       & 0.5647                                       & 0.4275                                         \\ 
 \bottomrule
\end{tabular}}
\end{center}
\end{table}

In Table~\ref{table:nn_lpips}, we report the average LPIPS score between the generated samples from our model to their closest real samples ranked by LPIPS score. In comparison, we show the baseline as the LPIPS between real images and their randomly augmented variants (randomly horizontal flipping and random cropping with $0.8$ spatial portion). We run each experiment 3 times with 100 randomly synthesized samples or real images, and report the lowest one. The std among the trials are usually lower than $0.005$. This experiment shows that, instead of memorizing the real images in the training set, our model is able to perceive the features from the real images, and generate images that are different and novel, in terms of compositions, shapes, and color patterns. 
\clearpage

\section{Decoder result}

\begin{figure}[h]
\vspace{-2mm}
\centering
\includegraphics[width=\linewidth,height=1.3\linewidth]{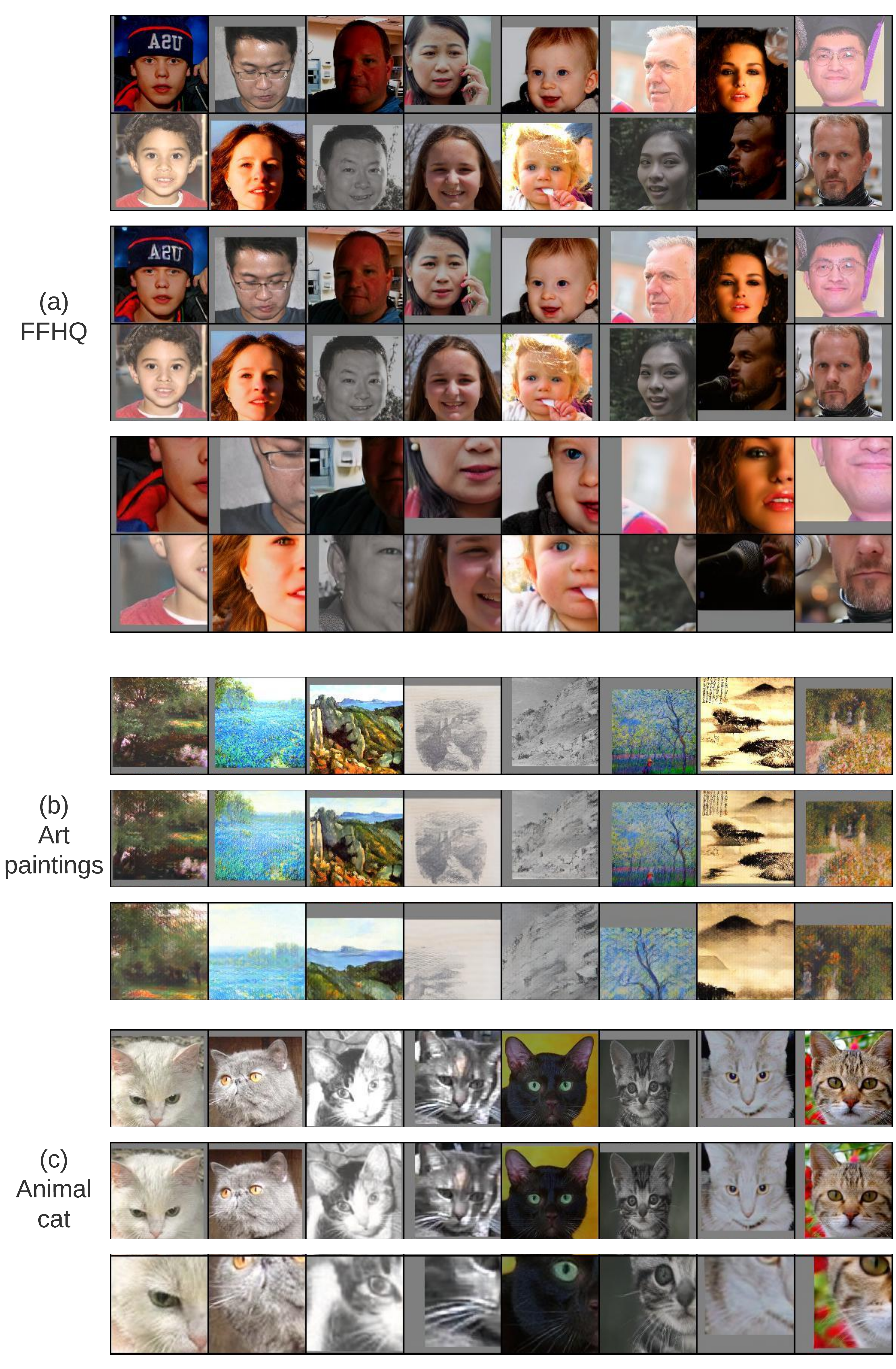}
\caption{\textbf{Reconstruction results from the decoder for training the auto-encoding discriminator}. For each dataset, the first panel shows the augmented real images during training, the second panel shows the reconstruction on full image, and the last panel shows the reconstruction on random cropped portions of the full image. Add the reconstructions are done on $128\times128$ resolution. }
\label{fig:decoder_recon}
\end{figure}


\end{document}